\documentclass[letterpaper]{article} % DO NOT CHANGE THIS
\usepackage{aaai24}
\usepackage{times}  % DO NOT CHANGE THIS
\usepackage{helvet}  % DO NOT CHANGE THIS
\usepackage{courier}  % DO NOT CHANGE THIS
\usepackage[hyphens]{url}  % DO NOT CHANGE THIS
\usepackage{graphicx} % DO NOT CHANGE THIS
\usepackage{natbib}  % DO NOT CHANGE THIS AND DO NOT ADD ANY OPTIONS TO IT
\usepackage{caption} % DO NOT CHANGE THIS AND DO NOT ADD ANY OPTIONS TO IT
\usepackage{algorithm}
\usepackage{algorithmic}
\usepackage{amsmath,amsfonts,bm}

\def\eqref#1{equation~\ref{#1}}
\def\1{\bm{1}}

\DeclareMathAlphabet{\mathsfit}{\encodingdefault}{\sfdefault}{m}{sl}
\SetMathAlphabet{\mathsfit}{bold}{\encodingdefault}{\sfdefault}{bx}{n}

\usepackage{newfloat}
\usepackage{listings}
\usepackage{enumitem}
\usepackage{booktabs,tabularx, multirow}
\DeclareCaptionStyle{ruled}{labelfont=normalfont,labelsep=colon,strut=off} % DO NOT CHANGE THIS
\floatstyle{ruled}
\newfloat{listing}{tb}{lst}{}
\floatname{listing}{Listing}

\usepackage{xcolor}

\title{UniAP: Towards Universal Animal Perception in Vision via Few-shot Learning}
\author{
    Meiqi Sun\textsuperscript{\rm 1}\equalcontrib \quad
    Zhonghan Zhao\textsuperscript{\rm 1}\equalcontrib \quad
    Wenhao Chai\textsuperscript{\rm 2}\equalcontrib \quad
    Hanjun Luo\textsuperscript{\rm 1} \\
    Shidong Cao\textsuperscript{\rm 1} \quad
    Yanting Zhang\textsuperscript{\rm 3} \quad
    Jenq-Neng Hwang\textsuperscript{\rm 2} \quad
    Gaoang Wang\textsuperscript{\rm 1\thanks{Corresponding author.}}
}
\affiliations{
    \textsuperscript{\rm 1}Zhejiang University \quad
    \textsuperscript{\rm 2}University of Washington \quad
    \textsuperscript{\rm 3}Donghua University
}

\begin{document}

\maketitle

\begin{abstract}
    Animal visual perception is an important technique for automatically monitoring animal health, understanding animal behaviors, and assisting animal-related research. However, it is challenging to design a deep learning-based perception model that can freely adapt to different animals across various perception tasks, due to the varying poses of a large diversity of animals, lacking data on rare species, and the semantic inconsistency of different tasks. We introduce UniAP, a novel Universal Animal Perception model that leverages few-shot learning to enable cross-species perception among various visual tasks. Our proposed model takes support images and labels as prompt guidance for a query image. Images and labels are processed through a Transformer-based encoder and a lightweight label encoder, respectively. Then a matching module is designed for aggregating information between prompt guidance and the query image, followed by a multi-head label decoder to generate outputs for various tasks. By capitalizing on the shared visual characteristics among different animals and tasks, UniAP enables the transfer of knowledge from well-studied species to those with limited labeled data or even unseen species. We demonstrate the effectiveness of UniAP through comprehensive experiments in pose estimation, segmentation, and classification tasks on diverse animal species, showcasing its ability to generalize and adapt to new classes with minimal labeled examples.
\end{abstract}
\begin{figure}[t]
    \centering
    \includegraphics[width=\linewidth]{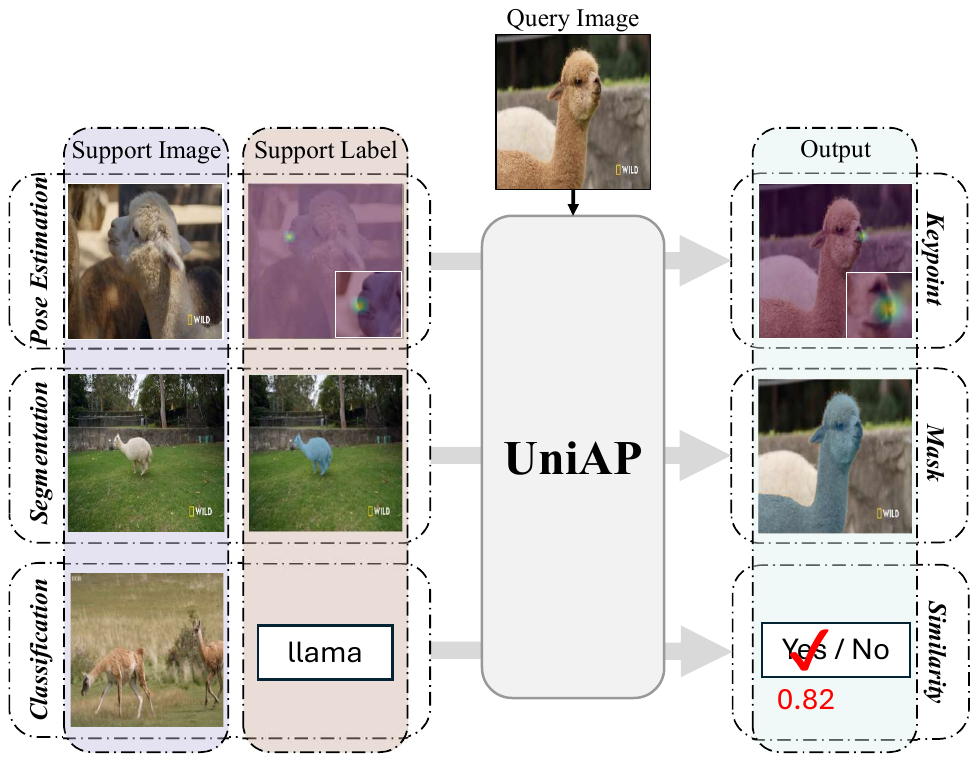}
    \caption{\textbf{UniAP} unifies animal pose estimation (top), segmentation (middle), and classification (bottom) tasks under a single model via few-shot learning. We use different support images of alpaca in Animal Kingdom dataset~\cite{ng2022animal} as examples for different tasks. 
    % In practice, different tasks can also utilize the same support image.
    }
    \label{fig:intro}
\end{figure}
\section{Introduction}

Animal perception plays an essential role in learning and understanding animal behaviors ~\cite{ashwood2022dynamic,anderson1990animal,butail2015fish}. Animal perception techniques have been employed in many visual tasks, such as pose estimation~\cite{li2023scarcenet,graving2019deepposekit}, pose tracking~\cite{pereira2022sleap, lauer2022multi}, face recognition~\cite{shi2020deep}, and semantic segmentation~\cite{li2021marine}.
Existing animal perception models usually focus on specific animal species on specific tasks.
To facilitate analyzing animal behaviors and assist animal-related research, it is highly demanded to design a universal animal perception model for diverse species and poses on various tasks. 
% Animal perception is an increasingly popular work, such as pose estimation~\cite{li2023scarcenet,graving2019deepposekit}, pose tracking~\cite{pereira2022sleap, lauer2022multi}, face recognition~\cite{shi2020deep}, and semantic segmentation~\cite{li2021marine}. Those perception models play an essential role in learning and understanding animal behavior~\cite{ashwood2022dynamic,anderson1990animal,butail2015fish}.

Due to various animal species with different poses and semantic inconsistency of various visual tasks, it is challenging to build a universal animal perception model. 
Firstly, the variety of animal species requires the perception model to have a high generalization capability. The mainstream approaches for animal perception rely on large-scale, manually labeled data for supervised training, which is inefficient and suffers from annotation quality issues. Adapting the model to rare animals with only a few data examples or unseen species is difficult.
Secondly, due to the semantic inconsistency of various tasks, it remains difficult to integrate information across different animal-related tasks and species to create a comprehensive representation of animal perception. This lack of unified cross-task models in multi-modal animal perception presents a significant challenge.

We claim that a universal few-shot model of animal perception tasks requires meeting certain criteria. Firstly, the model should have a unified architecture that can handle multiple tasks and share parameters across tasks to acquire generalizable knowledge. Secondly, the model ought to have the ability to work effectively with various animal species, even in situations with limited data, allowing it to use few-shot prompts and yield superior outcomes across a wider range of species.

We propose \textbf{UniAP}, a \underline{Uni}versal \underline{A}nimal \underline{P}erception model that leverages few-shot learning to enable cross-species perception among various visual tasks, as shown in Fig.~\ref{fig:intro}.
Our proposed model takes prompt images and labels from different modalities, such as poses and masks, as the support set. With prompt guidance, UniAP can generate multi-task outputs for a query image. 
Specifically, images and labels are processed through a Transformer-based encoder with task-specific learnable bias parameters and a lightweight label encoder. Then a matching module is designed for aggregating information between prompt guidance and the query image, followed by a multi-head label decoder to generate outputs for various tasks.
By capitalizing on the shared visual characteristics among different animals and tasks, UniAP enables knowledge transfer from well-studied species to those with limited labeled data or unseen species. We demonstrate the effectiveness of UniAP through comprehensive experiments in pose estimation, segmentation, and classification tasks on diverse animal species, showcasing its ability to generalize and adapt to new classes with minimal labeled examples.

Our core contributions are summarized as follows:
\begin{itemize}
    \item We propose UniAP, a novel universal model that tackles the problem of unified learning of animal perception tasks: pose estimation, semantic segmentation, and classification.
    \item UniAP can take images and labels from various modalities as prompt guidance to generate outputs for rare species or even unseen animals with a few examples. 
    \item Extensive experiments and ablation studies on various tasks and datasets demonstrate the effectiveness of UniAP.
\end{itemize}

% \vspace{180pt}
\section{Related Works}

\subsection{Animal Pose Estimation}
The pose estimation task has achieved significant success in humans~\cite{andriluka20142d,munea2020progress,dang2019deep} or vehicles~\cite{fang2019intention,lopez2019vehicle}. There have been some prior works showing interest in pose estimation for animals~\cite{pereira2019fast,graving2019deepposekit,lauer2022multi,cao2019cross,li2021synthetic,zhang2022promptpose,jiang2022animal,shooter2021sydog}. Animal pose estimation plays an essential role in learning and understanding animal behavior~\cite{anderson1990animal,butail2015fish}. Recently, 
POMNet~\cite{xu2022pose} aims to create a pose estimation model capable of detecting the pose of any class of object given only a few samples with keypoint definition. 
ScarceNet~\cite{li2023scarcenet} tackles the task of animal pose estimation with the setting that only a small set of labeled data and unlabeled images are available.
CLAMP~\cite{zhang2023clamp} attempts to bridge the gap by adapting the text prompts to the animal keypoints.
FSKD~\cite{lu2022few} is the first attempt to model keypoint detection as few-shot learning.
There are also several benchmarks like APT-36k~\cite{yang2022apt} and AP-10k~\cite{yu2021ap} facilitating the research in animal pose estimation. 

\subsection{Animal Semantic Segmentation and Classification}
%yu2018bisenet
Semantic segmentation~\cite{guo2018review,wang2018understanding} is a fundamental task in computer vision that assigns a class label to each pixel in an image. Recently, some works extend it to open-vocabulary~\cite{luo2023segclip,ma2022open}, \textit{i.e.}, segmenting objects from any categories
by their textual names or description. SAM~\cite{kirillov2023segment} aims to return a valid segmentation mask given any prompt, which opens up the possibility of zero-shot general object segmentation. Some works also focus on animal semantic segmentation~\cite{li2021marine}. But there is no specific benchmark for animal semantic segmentation. As for animal classification, some widely used benchmarks like MSCOCO~\cite{lin2014microsoft} and ImageNet~\cite{russakovsky2015imagenet} have plenty of images for animal classification.

\subsection{Few-shot Learning in Computer Vision}
Few-shot learning is a fundamental paradigm in computer vision that carries the promise of alleviating the need for exhaustively labeled data and has been widely explored in vision tasks such as semantic segmentation~\cite{shaban2017one,iqbal2022msanet,fan2022self}, instance segmentation~\cite{michaelis2018one,fan2020fgn}, and object detection~\cite{fan2020few,kang2019few}. Recently,
VTM~\cite{kim2023universal} proposes a universal few-shot learner for arbitrary dense prediction tasks. It utilizes non-parametric matching on embedded tokens of images and labels at the patch level, encompassing all tasks.

% \subsection{Multi-task Learning in Computer Vision}
% Building a unified framework for various computer vision tasks attract considerable attention in recent years. Some works~\cite{zou2023generalized,wang2022ofa,wang2022image,li2022mplug} focus on combining language knowledge to solving multi-modality tasks~\cite{chai2022deep}. Traditional multi-task learning~\cite{zhang2018overview} focus on designing a network~\cite{misra2016cross,fan2022m3vit} for various computer vision tasks (such as image classification, object detection, and semantic segmentation, \textit{etc.}) or training strategies~\cite{kendall2018multi,chen2018gradnorm,liu2019end}. Previous works design a shared backbone and then different heads for various tasks. Recently, there have also been efforts to adapt transformer-based networks through query design~\cite{wang2023visionllm,ci2023unihcp}.
\definecolor{IE}{RGB}{21,96,130}
\definecolor{LE}{RGB}{160,43,147}
\definecolor{MM}{RGB}{15,158,213}
\definecolor{LD}{RGB}{78,167,46}
\definecolor{CH}{RGB}{25,107,36}
\begin{figure*}[t]
    \centering
    \includegraphics[width=0.95\linewidth]{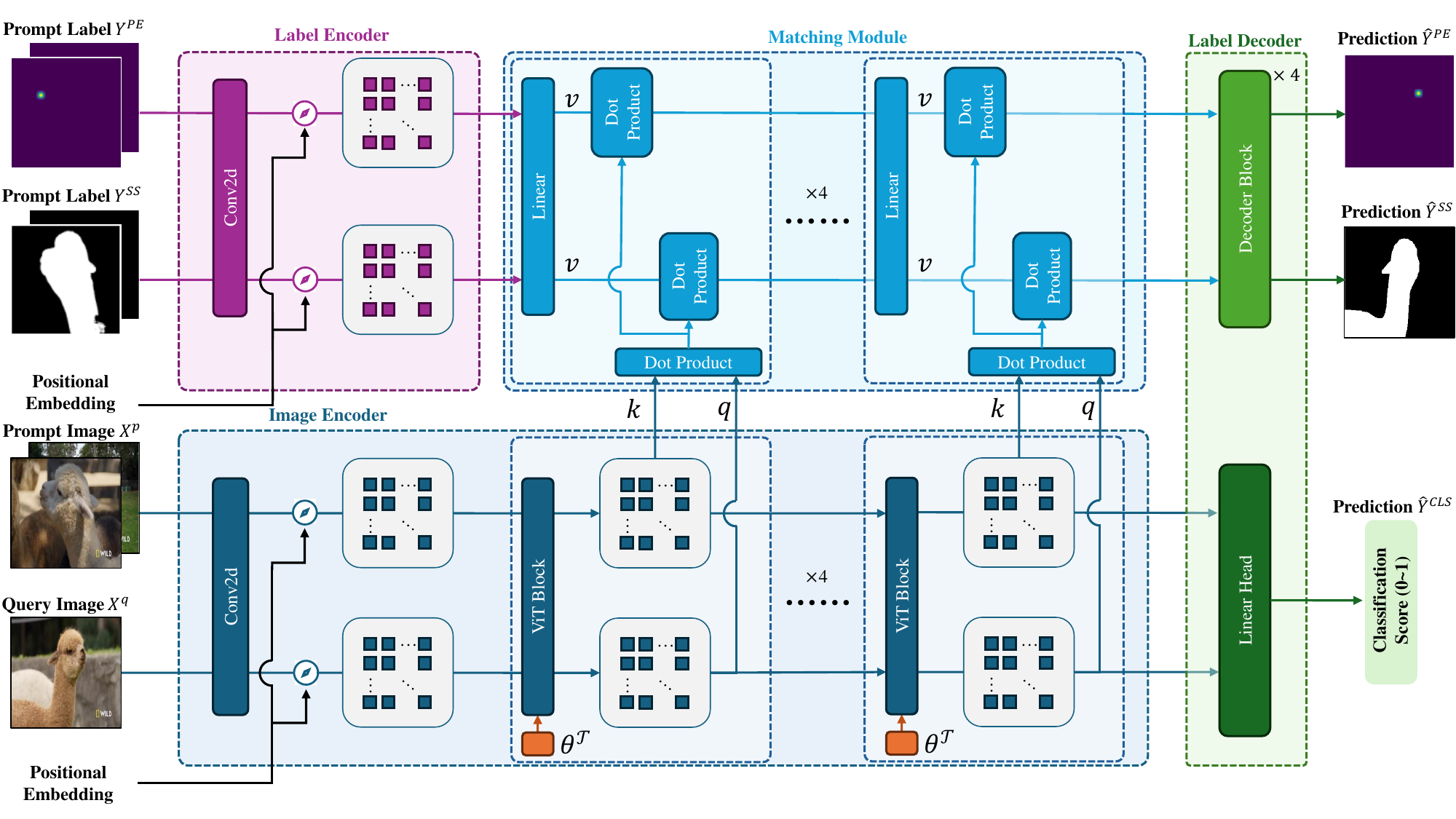}
    \caption*{
    \parbox[c][8pt][l]{8pt}{\colorbox{IE}{}}Image Encoder\quad
    \parbox[c][8pt][l]{8pt}{\colorbox{LE}{}}Label Encoder\quad
    \parbox[c][8pt][l]{8pt}{\colorbox{MM}{}}Matching Module\quad
    \parbox[c][8pt][l]{8pt}{\colorbox{LD}{}}Label Decoder\quad
    % \parbox[c][8pt][l]{8pt}{\colorbox{CH}{}}Classfication head
    }
    \caption{Overall architecture of UniAP.
    Our model is a hierarchical encoder-decoder with \textbf{Query Image} $X^{q}$ given only a few labeled examples \textbf{Prompt Set} $\mathcal{P} = \{(X^p, Y^p)\}$. 
    For any given task: pose estimation (PE) or semantic segmentation (SS), specific \textbf{Prompt Images} $X^{p}$ correspond to \textbf{Prompt Label} $Y^{PE}$ or $Y^{SS}$.
    Note that UniAP uses the feature pyramid network \textbf{(FPN)} to maintain the Affinity Matrix through multi-head attention layers in each hierarchy's matching module.}
    \label{fig:method}
\end{figure*}
\section{Methodology}

We propose \textbf{UniAP}, a \underline{Uni}versal \underline{A}nimal \underline{P}erception framework, which is designed to flexibly adapt to diverse animals across domains with few-shot examples for multiple unified tasks. The overall architecture of UniAP is shown in Fig.~\ref{fig:method}. We first present the overview of our proposed UniAP, and then introduce the architecture of the model, followed by the training and inference of the framework.

\subsection{Overview}
We develop a unified few-shot animal perception model denoted as $\mathcal{F}$, which can provide predictions $\hat{Y}^{q}$ on pose estimation (PE), semantic segmentation (SS) and classification (CLS) for an unseen image (\emph{query}) $X^{q}$ given only a few labeled examples (\emph{prompt set}) $\mathcal{P}$ for any given task in $\mathcal{T}=\{PE, SS\}$:
\begin{equation}\label{eqn:universal_few_shot_model}
    \hat{Y}^{q} = \mathcal{F}(X^{q}; \mathcal{P}), \quad \mathcal{P} = \{(X^p_i, Y^p_i)\}_{i\leq N},
\end{equation}
where $(X^p_i, Y^p_i)$ represents an image-label pair in the prompt set, and $N$ is the size of the prompt set.

We aim to find a universal function form $\mathcal{F}$ for Eq.~(\ref{eqn:universal_few_shot_model}) that can generate organized labels for any task within a unified framework. The query label is obtained by combining the prompt labels in a calculated manner.
Given a query image $X^q$ and a prompt set $\{(X^p_i, Y^p_i)\}_{i\leq N}$ with image-label pairs, the query label $Y^q$ can be obtained as follows,
% we project all images to embedding spaces and predict the query label $Y^q$ by,
\begin{equation}
\begin{split}
\mathcal{F}&\left(X^{q}; \mathcal{P}\right) = \\
&h\left(\mathcal{M}\left(f_\mathcal{T}\left(X^q\right), \{f_\mathcal{T}\left(X^p_i\right)\}_{i\leq N}, \{g\left(Y^p_i\right)\}_{i\leq N}\right)\right),
\end{split}
\end{equation}
where $f_\mathcal{T}$, $g$, $\mathcal{M}$ and $h$ are image encoder, label encoder, matching module, and label decoder, respectively. 

Specifically, prompt images $\{X^p_i\}_{i\leq N}$ and the query image $X^{q}$ are encoded by the image encoder $f_\mathcal{T}$.
Note that $f_\mathcal{T}$ contains task-specific lightweight trainable parameters $\theta_\mathcal{T}$.
In our approach, we incorporate adaptability with task-specific parameters $\theta_\mathcal{T}$, which is crucial in reflecting the unique features of each task. 
% To achieve this, we maintain lightweight task-specific parameters in the image encoder $f_\mathcal{T}$.
Simultaneously, prompt labels $\{Y^p_i\}_{i\leq N}$ are encoded by a lightweight label encoder $g$, where $g$ is shared across different tasks.
After encoding, the embedded features of prompt set and the query image are aggregated in the matching module $\mathcal{M}$. The matching module $\mathcal{M}$ is formulated as follows,
\begin{equation}
\label{eqn:matching}
\begin{split}
&\mathcal{M}\left(f_\mathcal{T}\left(X^q\right), \{f_\mathcal{T}\left(X^p_i\right)\}_{i\leq N}, \{g\left(Y^p_i\right)\}_{i\leq N}\right)
= \\
&\ \ \ \ \ \ \ \ \ \sum_{i\leq N} \sigma \left( f_\mathcal{T}(X^q), f_\mathcal{T}(X^p_i) \right) \cdot g(Y^p_i),
\end{split}
\end{equation} 
% where $h$ is the label encoder, $M$ is the matching module, and $\sigma$ is a similarity function that maps image patch embeddings to values between 0 and 1. $f_\mathcal{T}(X)=f(X;\theta_\mathcal{T},\theta_\mathcal{T})$ and $g(Y)=g(Y;\phi)$ correspond to the image and label encoder, respectively.
where $\sigma$ is a similarity function that maps image patch embeddings to values between $0$ and $1$. In the implementation, multi-head attention is employed in the matching module. We apply this matching process among all the tasks.
% In Equation~\ref{eqn:matching}, the matching module takes raw categorical labels $Y$ and interpolates them for classification. Then, we perform matching on the general embedding space of the label encoder $g(Y)$. Note that the matching process works consistently regardless of the task.
% In our approach, we incorporate adaptability by modulating the similarity with task-specific parameters $\theta_\mathcal{T}$, denoted by $\sigma(f_\mathcal{T}(X^q), f_\mathcal{T}(X^p))$, which is crucial in reflecting the unique features of each task. To achieve this, we maintain some task-specific parameters in the image encoder $f_\mathcal{T}$. 
Finally, the label decoder $h$ projects the embeddings to the results. 

Once trained, UniAP can easily adapt to \emph{unseen} animal images at test time.
% Since the label encoder $g$ is shared across tasks, we can use it to embed the label patches of unseen tasks with frozen parameters $\phi$.
% Adaptation to a novel task is performed by the image encoder $f_\mathcal{T}$, by optimizing the task-specific parameters $\theta_\mathcal{T}$, which take a small portion of the model.
This allows our model to adapt robustly to unseen animal images with a small prompt set.
In brief, UniAP has a unified architecture and shares most of the parameters across tasks that can acquire generalizable priors, improving its few-shot learning capabilities.

\subsection{Architecture}
\label{sec:arch}
Our model follows a hierarchical encoder-decoder architecture that implements similarity matching with four components: image encoder $f_\mathcal{T}$, label encoder $g$, the matching module $M$, and label decoder $h$.
Given the query image and the prompt set, the image encoder extracts patch-level embeddings (\emph{tokens}) of each query and prompt image independently.
The label encoder similarly extracts tokens of each prompt label.
Given the tokens at each hierarchy, the matching module performs matching to infer the tokens of the query label, from which the label decoder forms the raw query label.

\paragraph{Image Encoder}
Our image encoder relies on a Vision Transformer (ViT)~\cite{dosovitskiy2020image}. The ViT processes each query and prompt image separately but with shared weights, resulting in a tokenized representation of image patches at multiple levels. Following the approach in~\cite{ranftl2021vision}, we extract tokens from four intermediate ViT blocks to generate hierarchical features. To ensure that our system can learn general representations for a wide range of tasks, we initialize the parameters using pre-trained BEiT~\cite{bao2021beit}, which is self-supervised and less biased towards specific tasks.

For various tasks $\mathcal{T}$, we employ ``bias tuning"~\cite{cai2020tinytl, zaken2022bitfit}. This involves the sharing of weights $\theta$ across all tasks, but the use of unique biases $\theta_\mathcal{T}$ for each individual task. By implementing this approach, we can effectively adjust to meet the demands of different tasks $\mathcal{T}$, ensuring optimal performance and accuracy.

\paragraph{Label Encoder}
The label encoder $g$ is a lightweight function that utilizes only the patch embedding with a 2D convolution block and the positional embedding to extract tokens of each prompt label $Y^q$. Note that the label encoder $g$ is shared across tasks. 

\paragraph{Matching Module}
UniAP uses the feature pyramid network (FPN)~\cite{lin2017feature} to maintain the affinity matrix through multi-head attention layers in each hierarchical matching module. We retrieve the tokens from the intermediate layers of image encoders for the query image $X^q$ as $\{\mathbf{q}_j\}_{j\leq M}$ and prompt images $X^p_i$ as $\{\mathbf{k}_{i \times j}\}_{i\leq N,j\leq M}$. Additionally, we obtain the tokens of prompt labels $Y^p_i$ as $\{\mathbf{v}_{i \times j}\}_{i\leq N,j\leq M}$. Then, we organize the tokens into row vectors. Next, we utilize a multi-head attention layer to deduce the query label tokens at the hierarchy in the following manner:
\begin{equation}
\label{eqn:multihead_attention}
    \text{MHA}(\mathbf{q},\mathbf{k},\mathbf{v}) = \text{Concat}(\mathbf{o}_1, ..., \mathbf{o}_H)w^O,
\end{equation}
where
\begin{equation}
    \mathbf{o}_h = \text{Softmax}\left(\frac{\mathbf{q}w_h^Q(\mathbf{k}w_h^K)^\top}{\sqrt{d_H}}\right)\mathbf{v}w_h^V, \label{eqn:single_head_attention}
\end{equation}
where $H$ is number of heads, $d_H$ is head size, and $w_h^Q,w_h^K,w_h^V\in\mathbb{R}^{d\times d_H}$, $w^O\in\mathbb{R}^{Hd_H\times d}$.

Note that each attention head in Eq.~(\ref{eqn:single_head_attention}) plays the role of matching with lightweight functions. This is achieved by calculating the similarity between the query and prompt image tokens $\mathbf{q}$ and $\mathbf{k}$, which determines the weight given to each prompt label token $\mathbf{v}$ in inferring the query label token. The similarity function used is the scaled dot-product attention. The multi-head attention layer in Eq.~(\ref{eqn:multihead_attention}) has several trainable projection matrices $w_h^Q,w_h^K,w_h^V$, which enables it to learn different branches (heads) of the matching algorithm, each with unique similarity functions.

\paragraph{Label Decoder}
The label decoder consists of a dense prediction head and binary classification head.
In the dense prediction head, to predict the query label of the original resolution, the dense prediction head combines query label tokens inferred at multiple hierarchies. We choose the multi-scale decoder architecture of Dense Prediction Transformer~\cite{ranftl2021vision} as it is compatible with ViT encoders and multi-level tokens. At each hierarchy of the decoder, the inferred query label tokens are first spatially concatenated to create a feature map of constant size.
% ($M\to h\times w$). 
Then, (transposed) convolution layers of varying strides are used to produce a feature pyramid of increasing resolution for each feature map. The multi-scale features are then progressively upsampled and fused by convolutional blocks, followed by a convolutional head for final prediction.
The dense prediction head shares all its parameters across tasks, enabling it to develop a versatile approach to decoding a structured label from predicted query label tokens. In addition, the dense prediction head output is single-channel, making it suitable for various tasks involving any number of channels.
The binary classification head takes the output of the image encoder directly, followed by a linear feed-forward layer to obtain the classification score output.

\begin{figure*}[!t]
    \centering
    \includegraphics[width=\linewidth]{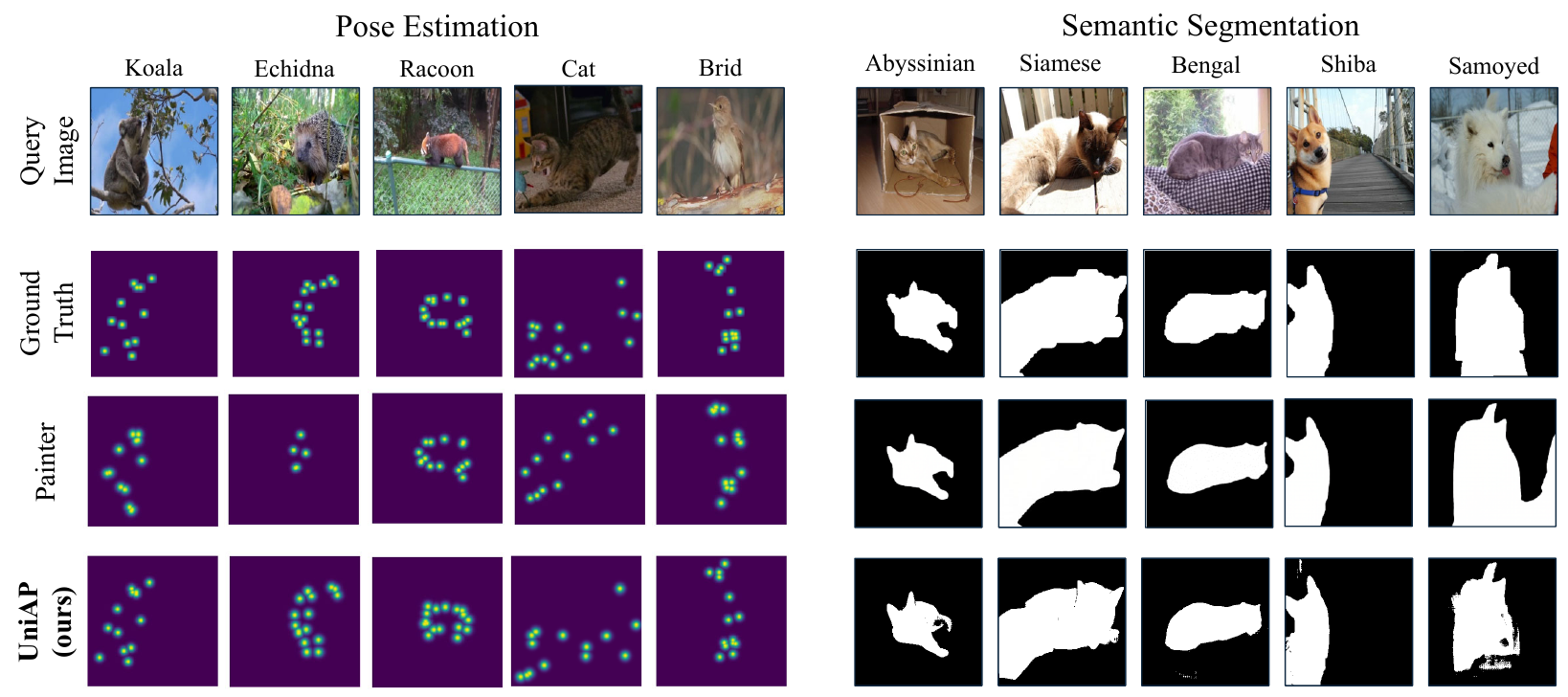}
    \caption{
    \textbf{Qualitative Comparison.}
    The comparison of representative methods in $1$/$0$-shot evaluation for pose estimation in the Animal Kingdom and semantic segmentation in the Oxford-IIIT dataset.
    }
    \label{fig:demo}
\end{figure*}

\subsection{Training and Inference}\label{sec:learning_and_inference} 

In the experiment, each dataset is split into two separate subsets by animal classes, namely $\mathcal{D}_\text{train}$ and $\mathcal{D}_\text{test}$, which does not have overlapping species. 
We train our model on a labeled dataset $\mathcal{D}_\text{train}$ of tasks $\mathcal{T} = \{PE, SS, CLS\}$.
% Note that we have two different versions of models that we train. The first model is trained only on a single task from $\mathcal{T}$, with the training episode always being chosen. Its training process is more fundamental and will not be described later.
% The second model is trained universally on all tasks in $\mathcal{T}$, with the training episode based on all tasks. 
At each episode, we sample two labeled sets $\mathcal{P}_\text{train}, \mathcal{Q}_\text{train}$ from $\mathcal{D}_\text{train}$.
Then we train the model to predict labels in $\mathcal{Q}_\text{train}$ using $\mathcal{P}_\text{train}$ as a prompt set.
Throughout universal training, we expose the model to tasks like PE, SS, and CLS in the training data set $\mathcal{D}_\text{train}$. This allows the model to gain a comprehensive understanding of all the tasks and be capable of handling few-shot input data. 
Besides, we utilize a principled multi-task loss incorporating homoscedastic task uncertainty~\cite{kendall2018multi} to learn multiple classification and regression losses with varying quantities and units.

Let $\mathcal{F}(X^{q}; \mathcal{P})$ denote the prediction of the model on $X^{q}$ using the prompt set $\mathcal{P}$. We construct the homoscedastic task uncertainty through probabilistic modeling to determine the most efficient distribution of varying task weights $w_i$.
Then the model ($f_\mathcal{T}, g, \mathcal{M}, h$) is trained by the following learning objective in end-to-end:
% \frac{1}{|\mathcal{Q}_\text{train}|}
\begin{equation}
    \underset{f_\mathcal{T}, g, \mathcal{M}, h}{\text{min}} 
    \sum_{(X^{q}, Y^{q}) \in \mathcal{Q}_\text{train}} w_t \mathcal{L}_t\left(
    Y^{q}, \mathcal{F}(X^{q}; \mathcal{P}_{\text{train}}))
    \right) + \log w_t,
    \label{eqn:training_objective}
\end{equation}
Where the variable $\mathcal{L}$ represents the loss function, the parameter $w_t$ represents the amount of noise in the task $t \in \mathcal{T}$. As the noise level increases, the weight of the corresponding loss function $\mathcal{L}_t$ decreases. Conversely, as the noise level decreases, the weight of the objective increases. To prevent the noise level from becoming too high and ignoring the data, the final term in the objective acts as a regulator for the noise parameters.
In our experiments, we use the cross-entropy loss for all tasks. 
During training, we train separate sets of bias parameters for the image encoder $f_\mathcal{T}$ for each training task. In the pose estimation task, we ensure that each prompt label of a single image contains only one keypoint.

Once the model is trained on $\mathcal{D}_\text{train}$, we use it to evaluate the \emph{unseen} animal species data $\mathcal{D}_\text{test}$ with a given prompt set $\mathcal{P}_{\text{test}}$. We adapt the model by fine-tuning the bias parameters of the image encoder $f_\mathcal{T}$ using the prompt set $\mathcal{P}_\text{test}$. To do this, we simulate episodic meta-learning by randomly dividing the prompt set into a sub-prompt set $\tilde{\mathcal{P}}$ and a sub-query set $\tilde{\mathcal{Q}}$.$\mathcal{P}_\text{test}=\tilde{\mathcal{P}}\mathop{\dot{\cup}}\tilde{\mathcal{Q}}$.
\begin{equation}
    \underset{\theta_\mathcal{T}}{\text{min}}~ 
    \frac{1}{|\tilde{\mathcal{Q}}|} \sum_{(X^{q}, Y^{q}) \in \tilde{\mathcal{Q}}} \mathcal{L}\left(
    Y^{q}, \mathcal{F}(X^{q}; \tilde{\mathcal{P}})
    \right),
\end{equation}
where $\theta_\mathcal{T}$ denotes bias parameters of the image encoder $f_\mathcal{T}$.
As part of our implementation, we adjust the input projection of the label encoder $g$ and the output head of the label decoder $h$, which has been shown to improve performance based on empirical evidence. These adjustments only require a small portion of parameters to be fine-tuned, which helps to prevent over-fitting on the limited prompt set $\mathcal{P}_{\text{test}}$. Once the model has been fine-tuned, we evaluate its effectiveness by using it to predict the label of an unseen query image with the prompt set $\mathcal{P}_{\text{test}}$.
\begin{table*}[t]
\centering
\resizebox{0.85\linewidth}{!}{
\begin{tabular}{@{}l|c|cc|cc|cc@{}}
\toprule
\multirow{2}{*}{\bf Model} & \multirow{2}{*}{\# shots} & \multicolumn{2}{c|}{\bf Animal Kingdom} & \multicolumn{2}{c|}{\bf Animal Pose} & \multicolumn{2}{c}{\bf APT-36K} \\
\cmidrule(lr){3-4} \cmidrule(lr){5-6} \cmidrule(lr){7-8}
& & PCK@0.2 & PCK@0.05 & PCK@0.2 & PCK@0.05 & PCK@0.2 & PCK@0.05 \\
\midrule
HRNet$_\text{w48}$ & - & 90.49 & 62.04 & \textbf{90.47} & 75.91  & 91.65 & 66.26\\
POMNet & 1 & 59.97 &30.65 & 73.28 & 51.81  & 63.90 &38.52\\ 
Painter &1 & 70.52 & 48.34 & 77.86 & 53.85 & 74.11 & 51.39 \\
\midrule
\multirow{2}{*}{\bf UniAP~(ours)} & 1  &64.44 &34.73 &76.67 &47.31 &85.31 &61.72 \\
                            & 30 / 35 / 40 &\textbf{99.65} &\textbf{98.59}&90.10&\textbf{77.78}&\textbf{96.47} &\textbf{86.18} \\
\bottomrule
\end{tabular}
}
\caption{\textbf{Pose estimation} on Animal Kingdom, Animal Pose, and APT-36K datasets. We list the one-shot performance and the best performance with its number of shots on the bottom line for each dataset. Note that Painter can only use one-shot, while ours can use multi-shot to boost the performance.}
\label{tab:pose_1}
\end{table*}
\begin{table}[t]
\centering
\resizebox{0.7\linewidth}{!}{
\begin{tabular}{@{}l|c|cc@{}}
\toprule
\bf Model & \# shots & Acc. & mIoU \\
\midrule
$\text{SAM}_{\text{user}}$ & - & 92.06 & 88.99\\
Painter & 1 & 86.47 & 77.72\\
\midrule
\multirow{2}{*}{\bf UniAP~(ours)} & 1  & 97.08 & 93.38 \\
                                  & 10 & \textbf{97.11} & \textbf{94.27} \\
\bottomrule
\end{tabular}
}
\caption{\textbf{Segmentation} on Oxford-IIIT dataset. We list the one-shot performance and the best performance with its number of shots on the bottom line.}
\label{tab:seg1}
\end{table}
\section{Experiments}

\subsection{Experiment Setting}

\subsubsection{Implementation Details}
We conduct the experiment using $8\times$ NVIDIA RTX 3090. During the training process, we randomly pair image prompts and queries of the same animal category. We crop the objects of interest based on their bounding box and resize them to $224 \times 224$.
To train our model, we utilize the Adam optimizer~\cite{kingma2015adam} for 1K iterations for warmup and 200K iterations in total. Our learning rate schedule follows the \emph{poly}~\cite{liu2015parsenet} method with $0.9$ decay rate, with base learning rates of $10^{-5}$ for pre-trained parameters and $10^{-4}$ for parameters trained from scratch. The learning rate 
We stop the training based on the validation metric threshold. Our global batch size is 64, with different tasks randomly sampled for each batch. We also include prompt and query sets in each batch of a size of 5 for each.
We adopt $\text{BEiT}_{\text{Base}}$~\citep{bao2021beit} backbone as the image encoder, which is pre-trained on ImageNet-22k~\cite{deng2009imagenet}.

\subsubsection{Datasets}
We apply UniAP on three datasets for animal perception tasks: Animal Kingdom~\cite{ng2022animal}, Animal Pose~\cite{cao2019cross}, and APT-36K~\cite{yang2022apt} for pose estimation. 
These datasets provide us with images of up to 49 kinds of animals with keypoint annotations.
Animal Kingdom dataset is also used for the classification task.
Additionally, we utilize the Oxford-IIIT Pet dataset~\cite{6248092} for the segmentation task. The Oxford-IIIT Pet is a 37-category dataset with segmentation annotations. 
Note that the categories of the train/test/val set are mutually exclusive.
We split several animal classes from the raw datasets, 12 from Animal Kingdom, 2 from Animal Pose, 6 from APT-36K, 5 from Oxford-IIIT Pet.

\subsubsection{Compared Baselines}
% We compare our UniAP with two different models: task-specific and unified. \textbf{Few-shot learning baselines} do not have access to the test tasks $\mathcal{T}_\text{test}$ during training and are given only a few labeled images at the test time. 
As no prior few-shot methods are developed for universal prediction tasks, we adapt state-of-the-art few-shot segmentation methods to our setup. We choose three models, Painter~\cite{wang2023images}, POMNet~\cite{xu2022pose}, and HRNet-w48~\cite{wang2020deep}. Painter is comprehensive on multi-tasks, while both POMNet and HRNet-w48 are specialized in pose estimation.
In addition, we adopt Painter~\cite{wang2023images} and SAM~\cite{kirillov2023segment} for prompt learning comparison,
% We compare our method, UniAP, with two different prompt learning approaches. For this comparison, we selected two models, Painter~\cite{wang2023images} and SAM~\cite{kirillov2023segment}. 
where Painter uses in-context prompting, while SAM needs user-interaction prompting.
To validate the classification task, we compare our method (UniAP) with the \text{clip}-{\text{vit}}-{\text{large}}-{\text{patch14}}~\cite{radford2021learning} to perform zero-shot classification on the animal dataset.

% \paragraph{Implementation Details.}
% We adopt $\text{BEiT}_{\text{Base}}$~\citep{bao2021beit} backbone as the image encoder of UniAP, which is pre-trained on ImageNet-22k~\cite{deng2009imagenet}. We compare UniAP with the most closed prior arts (\textif{e.g.} Painter~\cite{wang2023images}, POMNet~\cite{xu2022pose}, HRNet~\cite{wang2020deep}, SAM~\cite{kirillov2023segment}, CLIP~\cite{radford2021learning}, and \textit{etc.}).
% Painter is pretrained on \text{ViT}-{\text{mae}}-{\text{large}}~\cite{he2022masked}. All input data is converted into images to meet the model's specifications.
% For POMNet, which relies on the convolutional encoder and it is nontrivial to transfer them to ViT, we use ResNet-50~\cite{he2016deep} backbone pre-trained on ImageNet-1k with image classification, which is their best-performing configuration.
% HRNet-w48 is a CNN-based supervised model well-known for its excellent performance in pose estimation. We have not made any modifications to this model.
% SAM is a foundation model that can be directly inferred without training. In SAM, users' clicks on the center of the foreground will be entered as the necessary prompt information.

\subsubsection{Evaluation Protocol}
For pose estimation, we follow the PCK (Probability of Correct Keypoint)~\cite{andriluka20142d} and mAP (mean average precision) protocol. In our experiments, we report the mAP and average PCK ($\sigma = 0.05 \text{ and } 0.2$) of all the categories in each split.
For semantic segmentation, we adopt pixel accuracy and binary segmentation protocols and report both the standard accuracy (Acc.) and mean intersection over union (mIoU) for all classes.
For classification, we utilize standard accuracy (Acc.).

\subsection{Main Results}

\paragraph{Pose Estimation}
As shown in Table~\ref{tab:pose_1}, UniAP achieves state-of-the-art performance when the optimal number of shots is selected by few-shot on pose estimation, which reaches to or exceeds the performance of fully supervised models, far ahead of prompt learning or in-context learning methods such as POMNet~\cite{xu2022pose} and Painter~\cite{wang2023images}. Note that Painter can only use one-shot, while UniAP can further use multi-shot to boost the performance.

\paragraph{Segmentation}
As shown in Table~\ref{tab:seg1}, UniAP completely surpasses the similar prompt learning method Painter~\cite{wang2023images} and even the foundation model SAM~\cite{kirillov2023segment} in semantic segmentation. 

\paragraph{Classification}
As shown in Table~\ref{tab:class}, UniAP completely exceeds the performance of the Foundation model CLIP~\cite{radford2021learning} with different parameter sizes.
% $\text{CLIP}_{\text{ViT-Base}}$ and $\text{CLIP}_{\text{ViT-Large}}$.

\paragraph{Qualitative Comparison}
As shown in Fig.~\ref{fig:demo}, the number of shots increases, the performance of the model increases as a whole, and there is a downward trend at the end. As the number of shots increases, the performance of the model increases as a whole, and there is a downward trend at the end. With the improvement of finetune model performance.
\definecolor{FT}{RGB}{237,125,49}
\definecolor{UN}{RGB}{91,155,213}
\begin{figure*}
    \centering
    \includegraphics[width=0.27\textwidth]{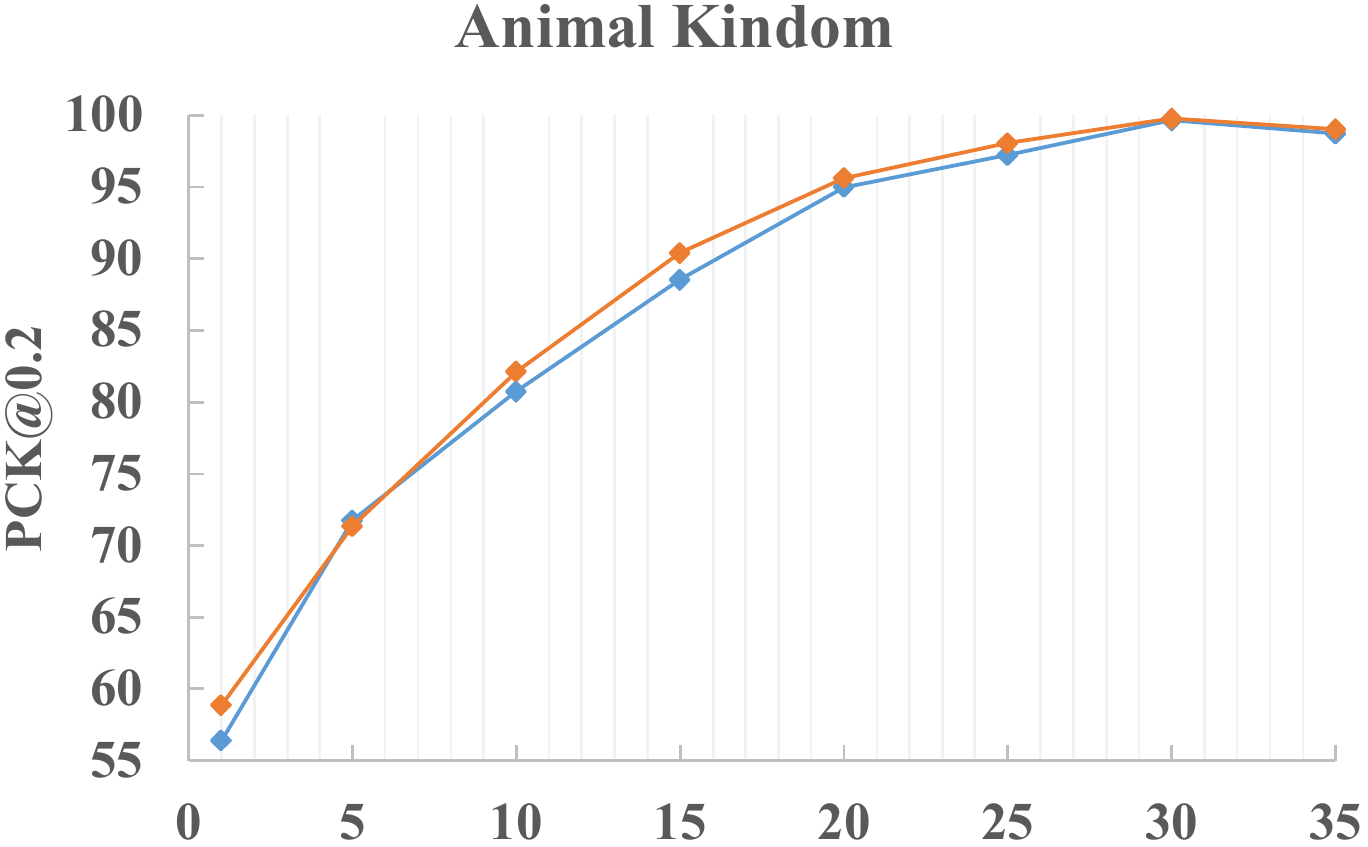}
    \includegraphics[width=0.27\textwidth]{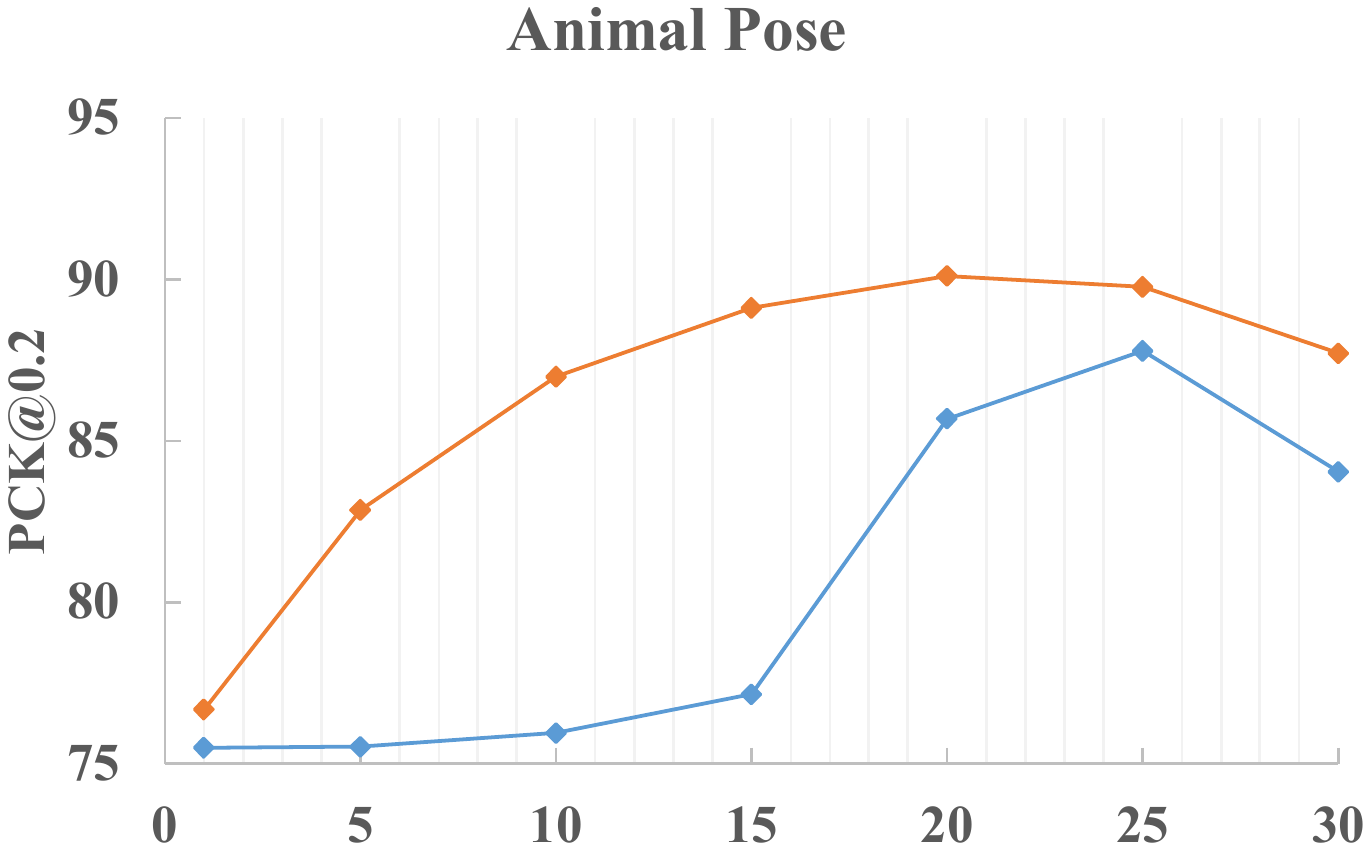}
    \includegraphics[width=0.27\textwidth]{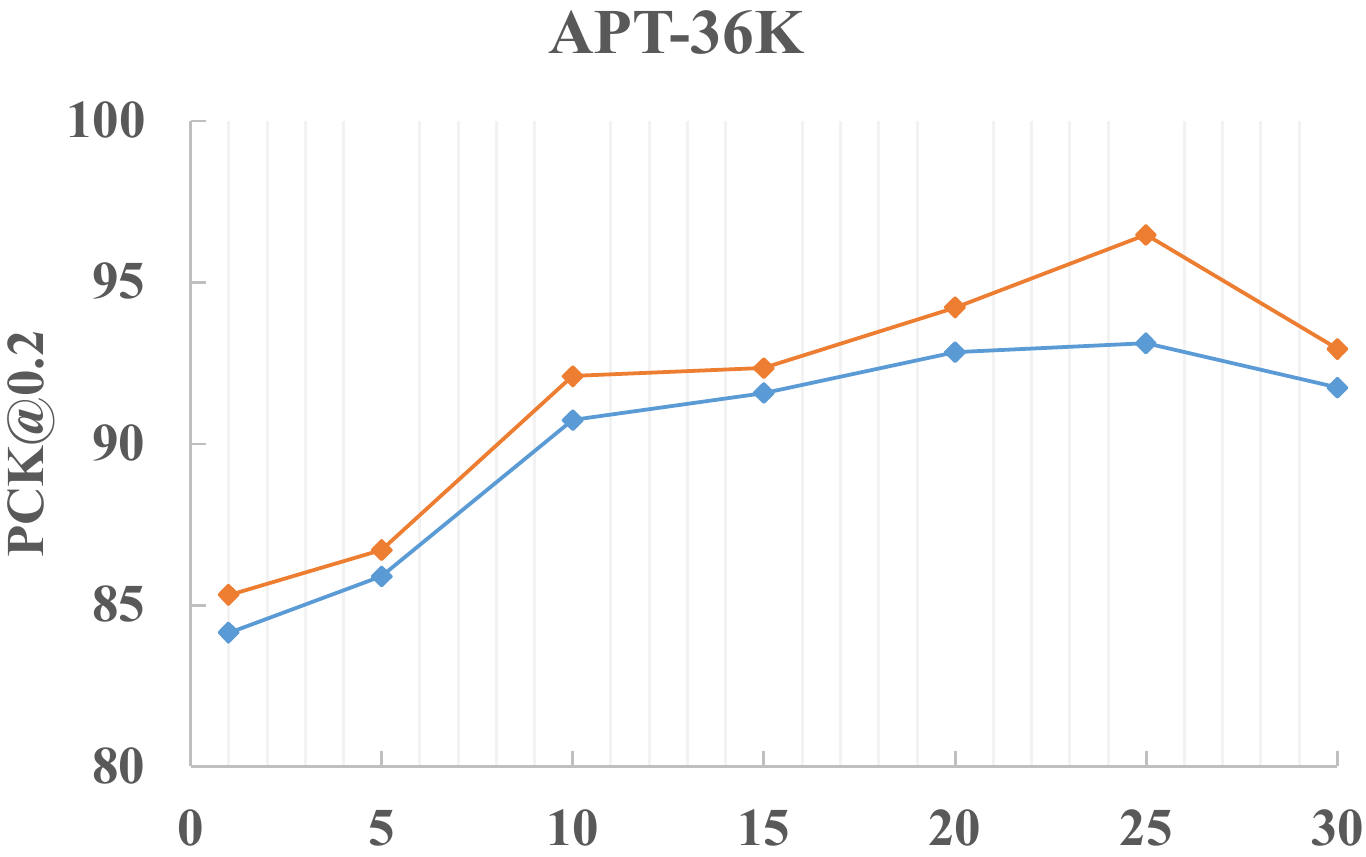}
    \\
    \includegraphics[width=0.27\textwidth]{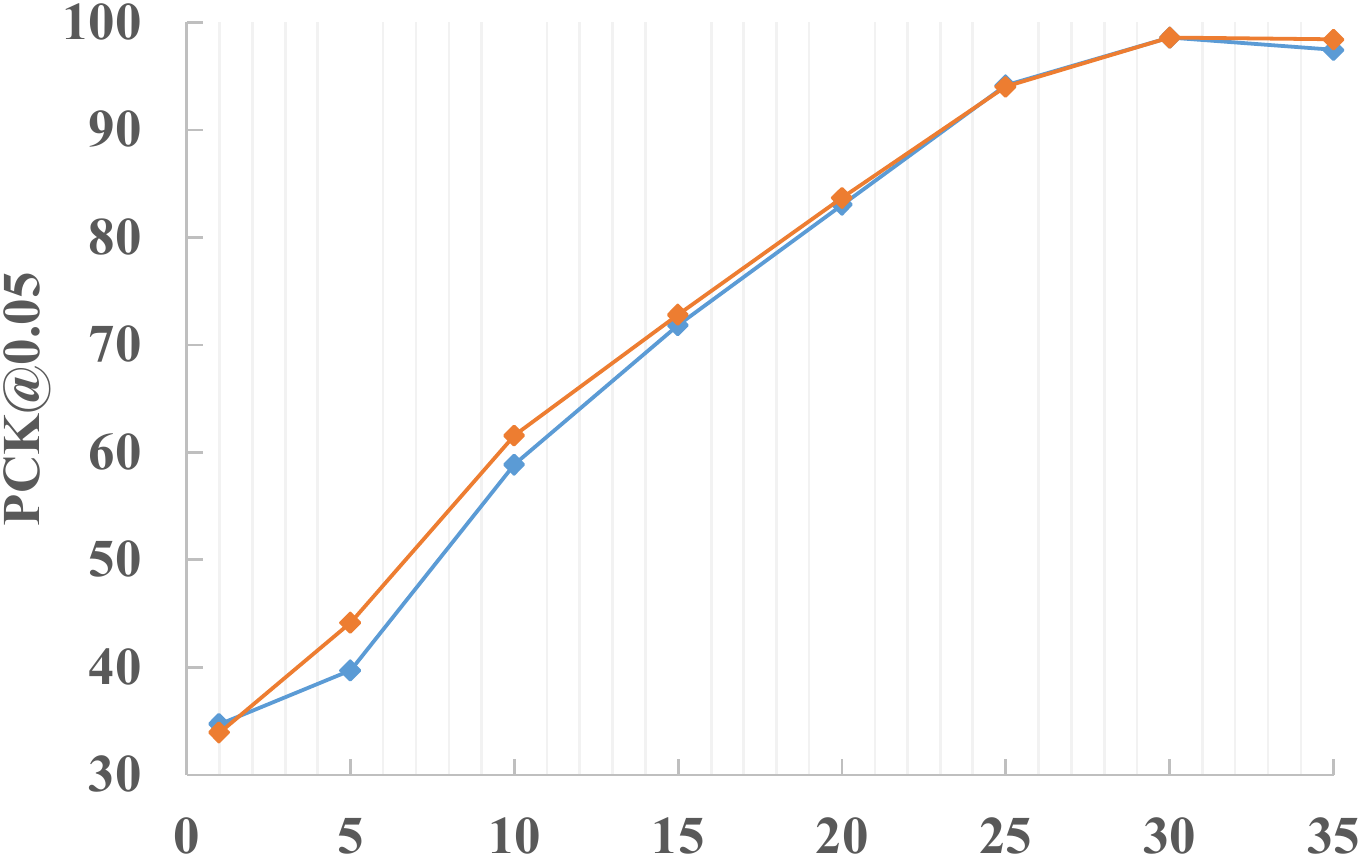}
    \includegraphics[width=0.27\textwidth]{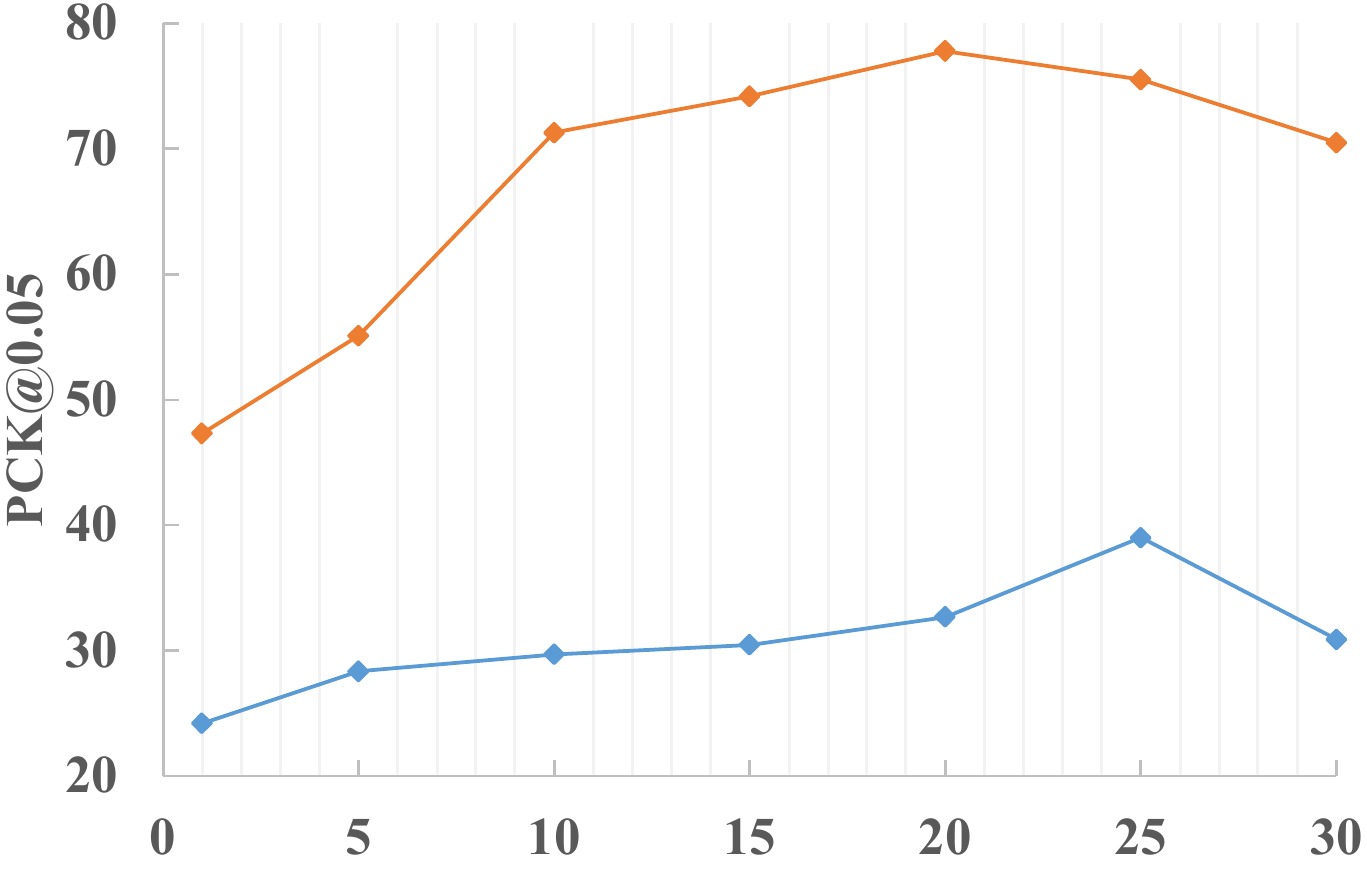}
    \includegraphics[width=0.27\textwidth]{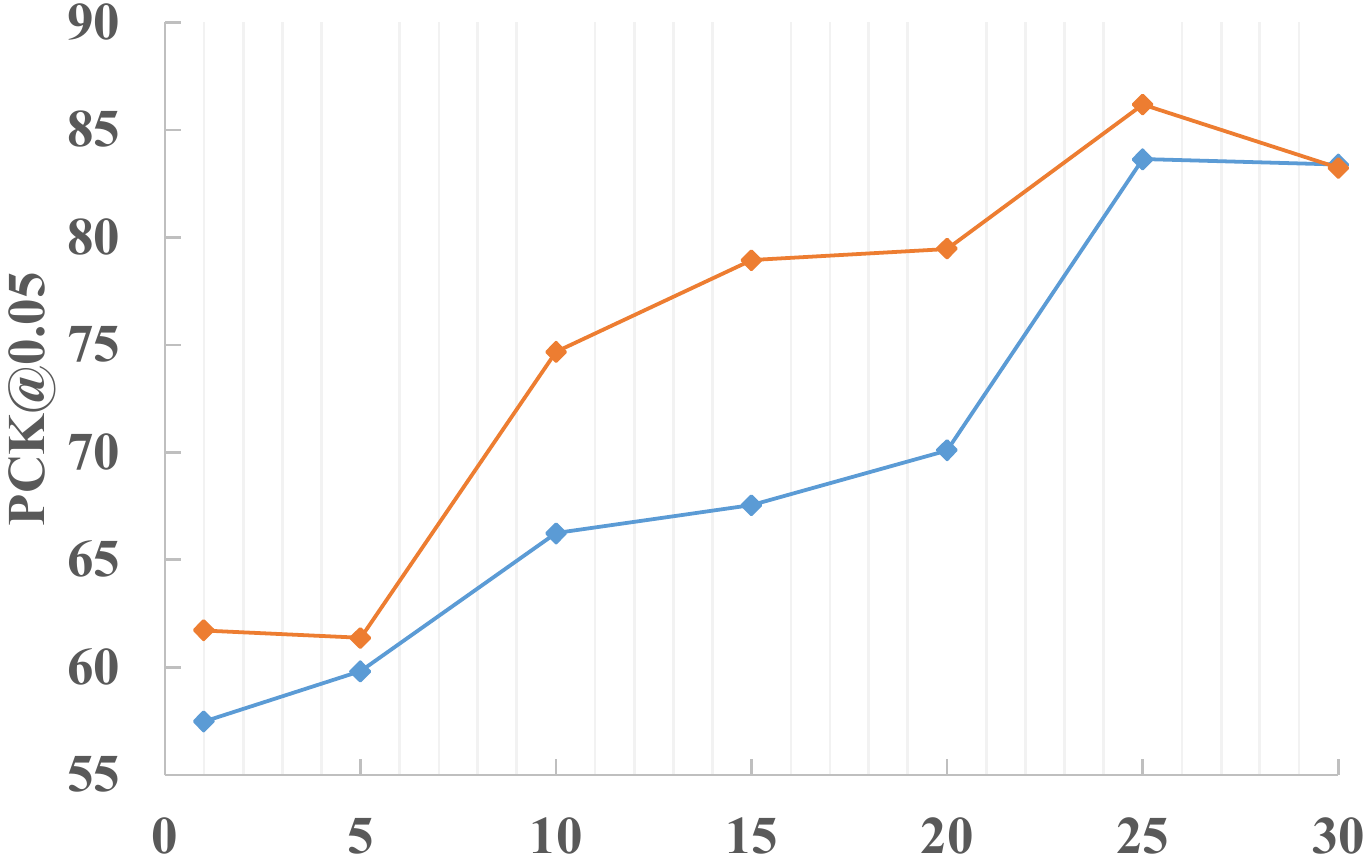}
    \caption*{
    \parbox[c][8pt][l]{8pt}{\colorbox{FT}{}}w. fine-tuning\quad
    \parbox[c][8pt][l]{8pt}{\colorbox{UN}{}}w.o. fine-tuning
    }
    \caption{
    \textbf{Ablation studies} on pose estimation performance with models w/o finetuning. According to the row, from top to bottom are PCK@0.2 and PCK@0.05, respectively. Note that the model performance is impacted by both the finetuning and the number of shots. The data quality is key to the initial performance and how it will perform over time. More shots mean receiving more prompt data, which improves its performance. The excessive number of shots will introduce additional noise and affect model performance. Thus, choosing the number of shots is a critical decision that can significantly impact the overall effectiveness of the model.
    }
    \label{fig:ablation}
\end{figure*}
\begin{table}[t]
\centering
\resizebox{0.55\linewidth}{!}{
\begin{tabular}{@{}l|c|c@{}}
\toprule
\bf Model & \# shots & Acc. \\
\midrule
$\text{CLIP}_{\text{ViT-Base}}$ & - & 88.54 \\
$\text{CLIP}_{\text{ViT-Large}}$ & - & 90.37 \\
\midrule
\multirow{2}{*}{\bf UniAP~(ours)} & 1 & 92.19\\
                                  & 5 & \textbf{93.13}\\
\bottomrule
\end{tabular}
}
\caption{\textbf{Classification} on Animal Kingdom dataset. We list the one-shot performance and the best performance with its number of shots on the bottom line.}
\label{tab:class}
\end{table}
\begin{table*}[!t]
\centering
\resizebox{0.85\linewidth}{!}{
\begin{tabular}{@{}l|c|cc|cc|cc@{}}
\toprule
\multirow{2}{*}{\bf Setting} & \multirow{2}{*}{\# shots} & \multicolumn{2}{c|}{\bf Animal Kingdom} & \multicolumn{2}{c|}{\bf Animal Pose} & \multicolumn{2}{c}{\bf APT-36K} \\
\cmidrule(lr){3-4} \cmidrule(lr){5-6} \cmidrule(lr){7-8}
& & PCK@0.2 & PCK@0.05 & PCK@0.2 & PCK@0.05 & PCK@0.2 & PCK@0.05 \\
\midrule
\multirow{2}{*}{\bf OOD} & 1 & 64.44 & 34.73 & 76.67 & 47.31 & 85.31 & 61.72 \\
              & 30 / 35 / 40 & \textbf{99.65} &\textbf{ 98.59} & 90.10 & 77.78 & 96.47 & 86.18 \\
\midrule
\multirow{2}{*}{\bf ID} & 1 & 77.39 & 61.25 & 77.69 & 46.74 & 92.41 & 61.41 \\
              & 20 / 20 / 35  & 99.26 & 98.10 & \textbf{94.97} & \textbf{88.47} & \textbf{96.92} & \textbf{92.70} \\ 
\midrule
\multirow{2}{*}{\bf CE} & 1 &54.18 &24.61 &60.47 & 28.47 &78.39 &47.92\\
                        & 5 / 5 / 5 &65.50 &28.86 &77.33 &50.62 &88.72 &71.67\\
\bottomrule
\end{tabular}
}
\caption{\textbf{Ablation studies} on evaluation settings. \textbf{ID}~(in domain) refers to the prompt and query sets the same in the category. \textbf{OOD}~(out of domain) and \textbf{CE}~(cross evaluation) refer to different categories of prompt and query sets, where in \textbf{CE} prompt sets are sourced from the training set, while all of the prompt sets in \textbf{OOD} are from the test set and have not been trained. We list the one-shot performance and the best performance with its number of shots on the bottom line of each part.}
\label{tab:domain}
\end{table*}
\subsection{Ablation Study}
In this section, we conduct extensive ablation studies to answer the following questions. 
% The main results are shown in Fig.~\ref{fig:ablation}.
\subsubsection{How does the number of shots matters?}
The results of our study can be seen in Fig.~\ref{fig:ablation}, where we demonstrate the effectiveness of UniAP. By gradually increasing the size of the prompt set from 1 to 40, our model reaches or outperforms both supervised methods and the foundation model across multiple tasks, even with a smaller amount of additional data. This suggests that UniAP could be particularly useful in specialized fields with limited labels.

\subsubsection{Do we actually need the task-specific fine-tuning stage?}
Through our observations in Fig.~\ref{fig:ablation}, we have noticed that combining multi-task training and few-shot adaptation with an efficient parameter-sharing strategy for bias tuning significantly improves performance with a noticeable gap.

\subsubsection{How does the domain affect performance?}
Based on the information provided in Table~\ref{tab:domain}, we have observed three different settings: In Domain (ID), Out of Domain (OOD), and Cross Evaluation (CE). 
To be specific, ID refers to the prompt and query sets the same in the category. OOD and CE refer to different categories of prompt and query sets, where in CE prompt sets are sourced from the training set, while all of the prompt sets in OOD are from the test set and have not been trained.
UniAP aims to maintain a consistent approach to animal perception tasks while also ensuring a comprehensive global understanding. This is achieved through task-specific fine-tuning. Additionally, we train on multiple large general-purpose animal datasets to ensure diversity. Our few-shot prompt mechanism allows for the efficient completion of tasks by utilizing prompt information.
\section{Limitation and Future Work}
Although UniAP employs a unified framework to handle animal pose estimation, segmentation, and classification tasks, several other tasks (such as object detection, behavior recognition, \textit{etc.}) remain to be addressed in future work. Besides, exploring these tasks in a video-based context (\textit{e.g.}, object tracking, video action recognition) is also worth investigating. Lastly, concerning animal perception, the quality and quantity of existing datasets are far from comparable to those of natural image datasets. The lack of fully annotated datasets hinders the sufficiency of our training. Therefore, collecting animal datasets or synthesizing through the virtual environment is a potential direction for future research. 
This would enable researchers to better understand the behavior of animals in their natural habitats.

\section{Conclusion}
This paper introduces UniAP, a universal animal perception vision model via few-shot learning, enabling the bridging of the gap between diverse animal species and visual tasks. Our proposed method integrating a transformer-based framework with task-specific bias tuning has demonstrated its efficacy in facilitating the cross-species animal perception learning process. Our proposed model takes support images and labels as prompt guidance for a query image. Images and labels are processed through a transformer-based encoder and a lightweight label encoder, respectively. Then the matching module aggregates information between prompt guidance and the query image, followed by a multi-head label decoder to generate outputs for various tasks. 

\bibliography{aaai24}

\newpage

\renewcommand{\thesection}{A}
\section{Ablation Studies}
This section provides additional results on the ablation experiments. We conducted experiments on pose estimation and semantic segmentation on 4 unique datasets (Animal Kingdom, Animal Pose, APT-36K, and Oxford-IIIT Pet), each comprising five distinct setups. These experiments validate the effectiveness of the multi-task loss and bias tuning strategy in multi-task learning. Note that we utilize a principled multi-task loss incorporating homoscedastic task uncertainty~\cite{kendall2018multi} to learn multiple classification and regression losses with varying quantities and units. To be more intuitive, the multi-task loss is mentioned as \textbf{awl}(adaptive weighted loss)  below. Meanwhile, the task-specific \textbf{bias tuning}~\cite{cai2020tinytl, zaken2022bitfit} is used in the image encoder for various tasks.

As shown in Fig.~\ref{fig:ablation_settings}, our method~(awl+bt) is basically close to the optimal~(awl+ft). Meanwhile, our method can save finetune time, as shown in Table~\ref{tab:number_of_tb}. The results show that using the awl strategy during training enhances performance across subtasks. Bias tuning during fine-tuning enables faster convergence while maintaining performance compared to full tuning.

\subsection{Adaptive loss schedule}
To get a better understanding of how our method achieves good generalization performance, we conducted an ablation study on the training procedure. Specifically, we used a 5-shot training approach.
We compare four models based on UniAP architecture with BEiT pre-trained encoder and different training procedures as follows.
\begin{itemize}[leftmargin=0.5cm]
    \item \textbf{baseline}: UniAP with multi-task trained with average weighted loss.
    \item \textbf{awl}: UniAP with multi-task trained with adaptive weighted loss.
    \item \textbf{bt}: UniAP with multi-task trained with task-specific bias tuning.
    \item \textbf{awl+bt} (Ours): UniAP with multi-task trained with adaptive weighted loss and task-specific bias tuning.
\end{itemize}

As shown in Table~\ref{tab:pose_AL_AK}, ~\ref{tab:pose_AL_AP}, ~\ref{tab:pose_AL_APT}, ~\ref{tab:seg_AL_OxP}, we observe that UniAP with naive 5-shot training (baseline) gets the lowest score, indicating a failure to adapt to the test examples in most of the tasks.
Then, we observe that UniAP with multi-task trained with adaptive weighted loss (awl) gets great performance, particularly on the Animal Kingdom and APT-36K data sets. There was also some improvement in the PCK@0.2 indicator on Animal Pose. Additionally, when UniAP was trained with task-specific bias tuning (bt), we noticed an improvement in performance across all datasets, especially on Animal Pose. 
In our method (awl+bt), we observe that multi-task training with awl loss for multi-task adaptation, combined with an efficient parameter-sharing strategy of bias tuning, further improves the performance with a clear gap with the baseline on each dataset which is close to or has reached the optimal performance (Animal Kingdom and Oxford-IIIT Pet even).
In conclusion, we can attribute Ours' fast generalization to the episodic training of various tasks, followed by parameter-efficient adaptation for multi-tasks.

\subsection{Bias tuning or full fine-tuning}

In order to get the ultimate performance, we replace the bias tuning of ours (awl+bt) with full tuning (awl+ft), \emph{i.e.}, finetune all parameters.

As shown in Table~\ref{tab:pose_AL_AK}, ~\ref{tab:pose_AL_AP}, ~\ref{tab:pose_AL_APT}, ~\ref{tab:seg_AL_OxP}, the performance of our method, awl+bt is basically close to that of awl+ft, and even exceeds it on Animal Kingdom and Oxford-IIIT Pet. As shown in Table~\ref{tab:number_of_tb}, our method can save finetune time.

\renewcommand{\thetable}{A}
\begin{table}[ht]
\caption{Number of tuning batches for bias tuning and full tunning.}
\label{tab:number_of_tb}
\begin{center}
    \resizebox{\linewidth}{!}{
    \begin{tabular}{c|cc}
        \toprule
        Dataset & bias tuning (awl+ft) & full tuning (awl+fft)\\
        \midrule
        Animal Kingdom  &3,119  &4,369    \\
        Animal Pose     &2,109  &2,565    \\
        APT-36K         &2,731  &4,005    \\
        \bottomrule
    \end{tabular}
}
\end{center}
\end{table}

\renewcommand{\thesection}{B}
\section{Additional Experiments}
\subsection{Other implementation details.}

% We adopt $\text{BEiT}_{\text{Base}}$~\cite{bao2021beit} backbone as the image encoder of UniAP, which is pre-trained on ImageNet-22k~\cite{deng2009imagenet}. We compare UniAP with the most closed prior arts (\textif{e.g.} Painter~\cite{wang2023images}, POMNet~\cite{xu2022pose}, HRNet~\cite{wang2020deep}, SAM~\cite{kirillov2023segment}, CLIP~\cite{radford2021learning}, and \textit{etc.}).

Painter is pretrained on \text{ViT}-{\text{mae}}-{\text{large}}~\cite{he2022masked}. All input data is converted into images to meet the model's specifications.
For POMNet, which relies on the convolutional encoder and it is nontrivial to transfer them to ViT, we use ResNet-50~\cite{he2016deep} backbone pre-trained on ImageNet-1k with image classification, which is their best-performing configuration.
HRNet-w48 is a CNN-based supervised model well-known for its excellent performance in pose estimation. We have not made any modifications to this model.
SAM is a foundation model that can be directly inferred without training. In SAM, users' clicks on the center of the foreground will be entered as the necessary prompt information.

\subsection{Parameter-Efficiency Analysis}
We conduct a comparison of the number of task-specific and shared parameters in our UniAP model and two prompt learning baselines, namely POMNet and Painter. Our aim was to determine the efficiency of our task adaptation.
POMNet is a single-task learning model that requires independent training for each new task, as no parameters are shared across tasks.
Painter, on the other hand, is a one-shot multi-task learning model that shares all parameters across tasks.

Our method is highly promising for the continual learning setting due to its extensive parameter-sharing. All task-specific knowledge is included in the bias parameters of the image encoder, allowing for the recall of past task knowledge without forgetting. This is achieved by keeping the corresponding bias parameters and switching to them whenever a past model is required.

    \renewcommand{\thetable}{B}
    \begin{table}[ht]
\caption{Number of task-specific and shared parameters for a task.}
\label{tab:number_of_parameters}
\begin{center}
\resizebox{\linewidth}{!}{
    \begin{tabular}{c|c|c|cc}
        \toprule
        Model & \# tasks & \# shots & Task-Specific & Shared \\
        \midrule
        POMNet & one & multiple & 58.25M & 0 \\
        Painter & multiple & one  & 0  & 353.55M \\
        Ours & multiple & multiple & 0.07M & 111.01M \\
        \bottomrule
    \end{tabular}
}
\end{center}
\end{table}
    
\renewcommand{\thesection}{C}
\section{Pseudo Code}

First, we use an image encoder to encode both the query image and prompt images as $\mathbf{q}$ and $\{\mathbf{k_i}\}_{i\leq N}$, respectively. Then, we use a label encoder to encode the prompt labels as $\{\mathbf{v_i}\}_{i\leq N}$. Next, as in the Eq.~\ref{eqn:matching}, we perform the matching module $\mathcal{M}$, which utilizes a multi-head attention layer and then the feature is decoded with the label decoder $h$ in order to predict the query label.

\begin{algorithm}
    \renewcommand{\algorithmicrequire}{\textbf{Input:}}
    \renewcommand{\algorithmicensure}{\textbf{Output:}}
    \caption{Universal Animal Perception (UniAP)}
    \label{alg:UniAP}
    \begin{algorithmic}[1]
        \REQUIRE Query Image $X_q$, Prompt Set $\mathcal{P}$
        \ENSURE Predictions $\hat{Y}^{q}$
        \STATE \textbf{Constants:} 
        \STATE \quad $\mathcal{T} = \{\text{Pose Estimation, Semantic Segmentation}\}$
        \STATE \textbf{Functions:}
        \STATE \quad $f_\mathcal{T}$: Image encoder with task-specific parameters $\theta_\mathcal{T}$
        \STATE \quad $g$: Label encoder (shared across tasks)
        \STATE \quad $\mathcal{M}$: Matching module
        \STATE \quad $h$: Label decoder
        \STATE \quad $\sigma$: Similarity function
        % $\hat{Y}^{q} = \mathcal{F}(X^{q}; \mathcal{P})$
        \STATE \textbf{Initialize:} 
        \STATE \quad $\mathbf{q} \gets f_\mathcal{T}(X_q)$
        \FOR{each $(X^p_i, Y^p_i)$ in $\mathcal{P}$}
            \STATE $\mathbf{k}_{i} \gets f_\mathcal{T}(X^p_i)$ \COMMENT{Encode using Image Encoder}
            \STATE $\mathbf{v}_{i} \gets g(Y^p_i)$ \COMMENT{Encode using Label Encoder}
        \ENDFOR
        \STATE $\mathbf{m} \gets \mathcal{M}(\mathbf{q}, \mathbf{k}_{i}, \mathbf{v}_{i})$
        \STATE $\hat{Y}^{q} \gets h(\mathbf{m})$ \COMMENT{Decode using Label Decoder}
        \RETURN $\hat{Y}^{q}$
    \end{algorithmic}
\end{algorithm}

\begin{algorithm}
    \renewcommand{\algorithmicrequire}{\textbf{Input:}}
    \renewcommand{\algorithmicensure}{\textbf{Output:}}
    \caption{Matching module $\mathcal{M}$ in UniAP}
    \label{alg:MatchingModule}
    \begin{algorithmic}[1]
        \REQUIRE $\mathbf{q}$, $\mathbf{k}$, $\mathbf{v}$
        \ENSURE $\mathbf{m}$
        \STATE \textbf{Constants:} 
        \STATE \quad Number of heads $H$, Head size $d_H$
        \STATE \quad Trainable projection matrices $w_h^Q,w_h^K,w_h^V$, $w^O$
        \FOR{$h = 1$ to $H$}
            \STATE $M_A \gets \frac{\mathbf{q}w_h^Q(\mathbf{k}w_h^K)^\top}{\sqrt{d_H}}$
            \STATE $\mathbf{o}_h \gets \text{Softmax}(M_A)\mathbf{v}w_h^V$
        \ENDFOR
        \STATE $\mathbf{m} \gets \text{Concat}(\mathbf{o}_1, ..., \mathbf{o}_H)w^O$
        \RETURN $\mathbf{m}$
    \end{algorithmic}
\end{algorithm}

\begin{equation}
\label{eqn:matching}
\begin{split}
&\mathcal{M}\left(f_\mathcal{T}\left(X^q\right), \{f_\mathcal{T}\left(X^p_i\right)\}_{i\leq N}, \{g\left(Y^p_i\right)\}_{i\leq N}\right)
= \\
&\ \ \ \ \ \ \ \ \ \sum_{i\leq N} \sigma \left( f_\mathcal{T}(X^q), f_\mathcal{T}(X^p_i) \right) \cdot g(Y^p_i),
\end{split}
\end{equation}
\clearpage

\renewcommand{\thefigure}{A}
\definecolor{1}{RGB}{83,125,201}
\definecolor{2}{RGB}{255,192,0}
\definecolor{3}{RGB}{146,208,80}
\definecolor{4}{RGB}{75,115,47}
\definecolor{5}{RGB}{221,90,9}

\begin{figure*}[t]
    \centering
    % \includegraphics[width=0.27\textwidth]{fig/supp/ak0.2.pdf}
    % \includegraphics[width=0.27\textwidth]{fig/supp/ap0.2.pdf}
    % \includegraphics[width=0.27\textwidth]{fig/supp/apt0.2.pdf}
    % \\
    % \includegraphics[width=0.27\textwidth]{fig/supp/ak0.05.pdf}
    % \includegraphics[width=0.27\textwidth]{fig/supp/ap0.05.pdf}
    % \includegraphics[width=0.27\textwidth]{fig/supp/apt0.05.pdf}
    \includegraphics[width=0.90\textwidth]{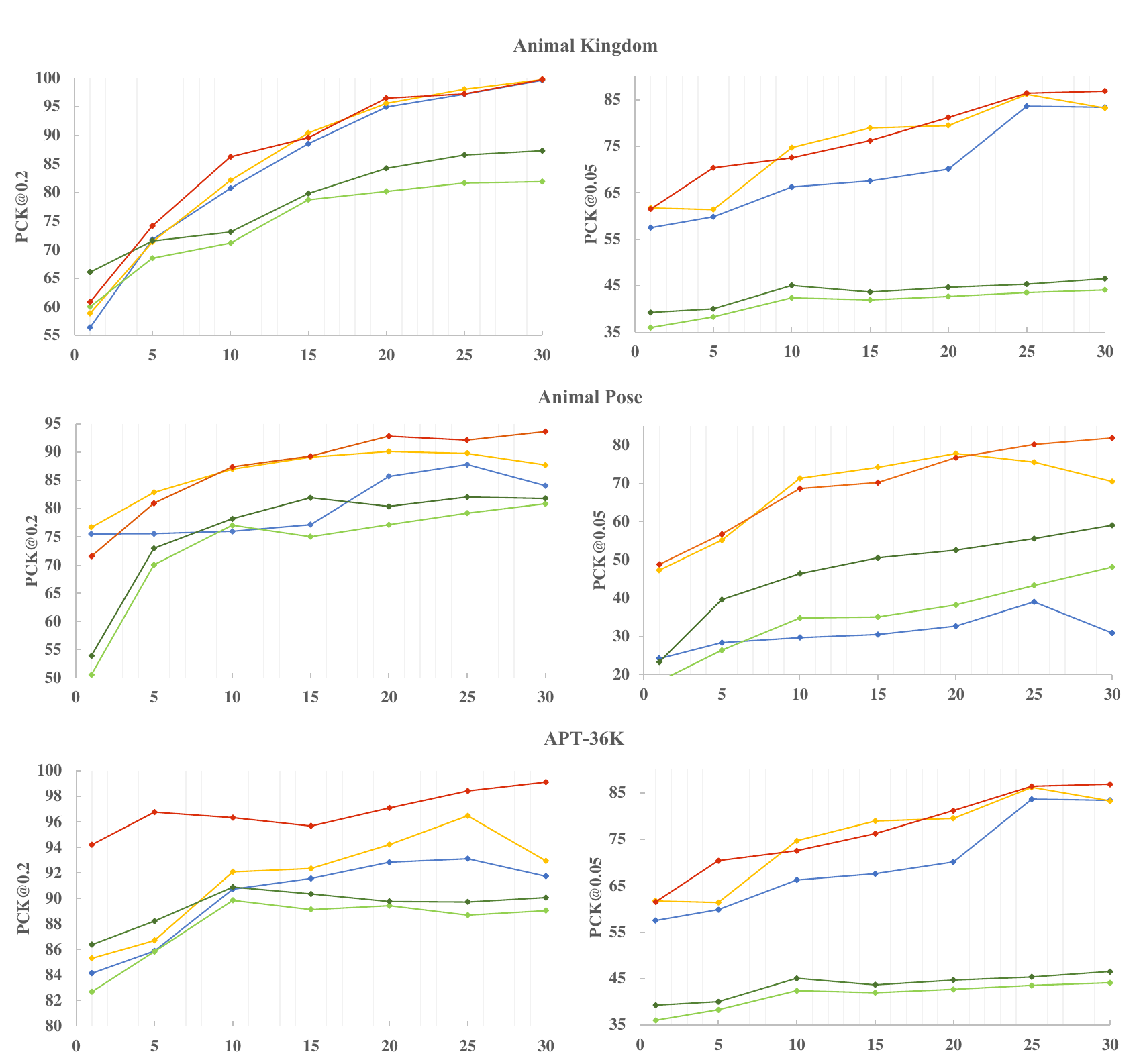}
    \caption*{
    \parbox[c][8pt][l]{8pt}{\colorbox{3}{}}baseline\quad
    \parbox[c][8pt][l]{8pt}{\colorbox{1}{}}awl\quad
    \parbox[c][8pt][l]{8pt}{\colorbox{4}{}}bt\quad
    \parbox[c][8pt][l]{8pt}{\colorbox{2}{}}awl+bt~(Ours)\quad
    \parbox[c][8pt][l]{8pt}{\colorbox{5}{}}awl+ft
    }
    \caption{
    \textbf{Ablation studies} on the performance of various shots with different settings. During experiments, the baseline refers to the raw model without any additional settings, "awl" refers to adaptive weighted loss, "bt" refers to bias tuning, and "ft" refers to full tuning.
    }
    \label{fig:ablation_settings}
\end{figure*}
\clearpage
\renewcommand{\thetable}{C}
\begin{table*}[t]
\centering
\resizebox{0.85\linewidth}{!}{
\begin{tabular}{c|cc|cc|cc|cc|cc@{}}
\toprule
\multirow{2}{*}{\# shots} & \multicolumn{2}{c|}{\bf baseline} & \multicolumn{2}{c|}{\bf awl} & \multicolumn{2}{c|}{\bf bt} &\multicolumn{2}{c|}{\bf awl+bt~(Ours)} &\multicolumn{2}{c}{\bf awl+ft}\\
\cmidrule(lr){2-3} \cmidrule(lr){4-5} \cmidrule(lr){6-7} \cmidrule(lr){8-9} \cmidrule(lr){10-11}
& PCK@0.2 & PCK@0.05 & PCK@0.2 & PCK@0.05 & PCK@0.2 & PCK@0.05 & PCK@0.2 & PCK@0.05 & PCK@0.2 & PCK@0.05\\
\midrule
1  &60.02  &19.17   &56.38 &34.73   &66.06  &22.65   &58.86 &33.93     &60.85 &33.04\\
5  &68.53  &28.74   &71.77 &39.74   &71.56  &30.23   &71.35 &44.18     &74.17 &44.96\\ 
10 &71.16  &31.24   &80.76 &58.89   &73.08  &33.20   &82.13 &61.60     &86.26 &70.46\\
15 &78.73  &38.18   &88.54 &71.87   &73.08  &40.11   &90.41 &72.85     &89.61 &77.31\\
20 &80.18  &41.37   &94.96 &83.03   &84.22  &57.46   &95.60 &83.67     &96.51 &88.62\\
25 &81.67  &43.58   &97.21 &94.14   &86.57  &63.14   &98.06 &94.04     &97.24 &94.03\\
30 &81.90  &48.14   &99.65 &98.59   &87.32  &66.26   &\textbf{99.76}   &\textbf{98.60}     &99.76 &98.59\\
\bottomrule
\end{tabular}
}
\caption{\textbf{Ablation studies of Animal Kingdom} on the performance of various shots with different settings. During experiments, the baseline refers to the raw model without any additional settings, "awl" refers to adaptive weighted loss, "bt" refers to bias tuning, and "ft" refers to full tuning.}
\label{tab:pose_AL_AK}
\end{table*}

\renewcommand{\thetable}{D}
\begin{table*}[t]
\centering
\resizebox{0.85\linewidth}{!}{
\begin{tabular}{c|cc|cc|cc|cc|cc@{}}
\toprule
\multirow{2}{*}{\# shots} & \multicolumn{2}{c|}{\bf baseline} & \multicolumn{2}{c|}{\bf awl} & \multicolumn{2}{c|}{\bf bt} &\multicolumn{2}{c|}{\bf awl+bt~(Ours)} &\multicolumn{2}{c}{\bf awl+ft}\\
\cmidrule(lr){2-3} \cmidrule(lr){4-5} \cmidrule(lr){6-7} \cmidrule(lr){8-9} \cmidrule(lr){10-11}
& PCK@0.2 & PCK@0.05 & PCK@0.2 & PCK@0.05 & PCK@0.2 & PCK@0.05 & PCK@0.2 & PCK@0.05 & PCK@0.2 & PCK@0.05\\
\midrule
1  &50.58  &18.30   &75.49 &24.19   &53.94  &23.32   &76.67 &47.31    &71.54 &48.81\\
5  &70.02  &26.39   &75.53 &28.37   &72.95  &39.60   &82.86 &55.11    &80.93 &56.72\\ 
10 &77.03  &34.78   &75.95 &29.71   &78.18  &46.41   &86.99 &71.30    &87.40 &68.60\\
15 &75.01  &35.11   &77.15 &30.47   &81.91  &50.55   &89.12 &74.20    &89.30 &70.17\\
20 &77.12  &38.21   &85.68 &32.66   &80.36  &52.55   &\textbf{90.10}  &\textbf{77.78}    &92.80 &76.72\\
25 &79.19  &43.32   &87.79 &39.02   &82.03  &55.54   &89.77 &75.52    &92.11 &80.12\\
30 &80.85  &48.12   &84.05 &30.92   &81.79  &59.04   &87.72 &70.48    &93.62 &81.83\\
\bottomrule
\end{tabular}
}
\caption{\textbf{Ablation studies of Animal Pose} on the performance of various shots with different settings. During experiments, the baseline refers to the raw model without any additional settings, "awl" refers to adaptive weighted loss, "bt" refers to bias tuning, and "ft" refers to full tuning.}
\label{tab:pose_AL_AP}
\end{table*}
\renewcommand{\thetable}{E}
\begin{table*}[t]
\centering
\resizebox{0.85\linewidth}{!}{
\begin{tabular}{c|cc|cc|cc|cc|cc@{}}
\toprule
\multirow{2}{*}{\# shots} & \multicolumn{2}{c|}{\bf baseline} & \multicolumn{2}{c|}{\bf awl} & \multicolumn{2}{c|}{\bf bt} &\multicolumn{2}{c|}{\bf awl+bt~(Ours)} &\multicolumn{2}{c}{\bf awl+ft}\\
\cmidrule(lr){2-3} \cmidrule(lr){4-5} \cmidrule(lr){6-7} \cmidrule(lr){8-9} \cmidrule(lr){10-11}
& PCK@0.2 & PCK@0.05 & PCK@0.2 & PCK@0.05 & PCK@0.2 & PCK@0.05 & PCK@0.2 & PCK@0.05 & PCK@0.2 & PCK@0.05\\
\midrule
1  &82.69  &36.01   &84.14 &57.49   &86.38  &39.27   &85.31 &61.72     &94.21 &61.50\\
5  &85.84  &38.29   &85.89 &59.83   &88.22  &40.03   &86.71 &61.38     &96.76 &70.37\\ 
10 &89.85  &42.41   &90.73 &66.24   &90.89  &45.05   &92.09 &74.69     &96.32 &72.54\\
15 &89.13  &41.98   &91.57 &67.55   &90.35  &43.66   &92.34 &78.94     &95.68 &76.23\\
20 &89.43  &42.69   &92.83 &70.09   &89.77  &44.69   &94.21 &79.47     &97.09 &81.17\\
25 &88.69  &43.52   &93.11 &83.64   &89.73  &45.36   &\textbf{96.47}   &\textbf{86.18}     &98.42 &86.42\\
30 &89.05  &44.10   &91.74 &83.39   &90.06  &46.53   &92.94 &83.23     &98.42 &86.86\\
\bottomrule
\end{tabular}
}
\caption{\textbf{Ablation studies of APT-36K} on the performance of various shots with different settings. During experiments, the baseline refers to the raw model without any additional settings, "awl" refers to adaptive weighted loss, "bt" refers to bias tuning, and "ft" refers to full tuning.}
\label{tab:pose_AL_APT}
\end{table*}
\renewcommand{\thetable}{F}
\begin{table*}[t]
\centering
\resizebox{0.6\linewidth}{!}{
\begin{tabular}{c|cc|cc|cc|cc|cc@{}}
\toprule
\multirow{2}{*}{\# shots} & \multicolumn{2}{c|}{\bf baseline} & \multicolumn{2}{c|}{\bf awl} & \multicolumn{2}{c|}{\bf bt} &\multicolumn{2}{c|}{\bf awl+bt~(Ours)} &\multicolumn{2}{c}{\bf awl+ft}\\
\cmidrule(lr){2-3} \cmidrule(lr){4-5} \cmidrule(lr){6-7} \cmidrule(lr){8-9} \cmidrule(lr){10-11}
& Acc. & mIoU & Acc. & mIoU & Acc. & mIoU & Acc. & mIoU & Acc. & mIoU\\
\midrule
1  &92.68  &89.06   &94.14 &92.49   &93.28  &92.27   &97.08 &93.38     &96.06 &92.35\\
5  &92.84  &90.29   &95.89 &92.83   &94.42  &92.54   &97.11 &93.41     &96.55 &93.38\\ 
10 &93.45  &91.41   &96.23 &93.24   &95.80  &92.75   &\textbf{97.11}   &\textbf{94.27}     &96.98 &94.15\\
\bottomrule
\end{tabular}
}
\caption{\textbf{Ablation studies of Oxford-IIIT Pet} on the performance of various shots with different settings. During experiments, the baseline refers to the raw model without any additional settings, "awl" refers to adaptive weighted loss, "bt" refers to bias tuning, and "ft" refers to full tuning.}
\label{tab:seg_AL_OxP}
\end{table*}

\end{document}